\documentclass[10pt,twocolumn,letterpaper]{article}

\usepackage{iccv}
\usepackage{times}
\usepackage{epsfig}
\usepackage{graphicx}
\usepackage{amsmath}
\usepackage{amssymb}

\usepackage{pifont}
\usepackage{algorithm}
\usepackage{algpseudocode}
\usepackage{multirow}
\usepackage{eso-pic}
\usepackage{booktabs}

\usepackage{array}
\newcommand{\PreserveBackslash}[1]{\let\temp=\\#1\let\\=\temp}
\newcolumntype{C}[1]{>{\PreserveBackslash\centering}p{#1}}
\newcolumntype{R}[1]{>{\PreserveBackslash\raggedleft}p{#1}}
\newcolumntype{L}[1]{>{\PreserveBackslash\raggedright}p{#1}}

\usepackage[pagebackref=true,breaklinks=true,letterpaper=true,colorlinks,bookmarks=false]{hyperref}

\iccvfinalcopy 

\def\iccvPaperID{5691} 

\ificcvfinal\fi

\begin{document}

\title{DRINet: A Dual-Representation Iterative Learning Network \\ for Point Cloud Segmentation}

\author{
	\begin{tabular}{ p{2.8cm}<{\centering} p{2.8cm}<{\centering} p{2.8cm}<{\centering} p{2.8cm}<{\centering}}
Maosheng Ye\textsuperscript{1$\ddagger$*} & Shuangjie Xu\textsuperscript{2*} & Tongyi Cao\textsuperscript{2}  & Qifeng Chen\textsuperscript{1} 
\end{tabular}\\
\textsuperscript{1}Hong Kong University of Science and Technology  \quad  \textsuperscript{2}DEEPROUTE.AI\\
{\tt\small myeag@connect.ust.hk \quad \{shuangjiexu, tongyicao\}@deeproute.ai \quad cqf@ust.hk}
}

\maketitle
\ificcvfinal\thispagestyle{empty}\fi

\begin{abstract}
We present a novel and flexible architecture for point cloud segmentation with dual-representation iterative learning. 
In point cloud processing, different representations have their own pros and cons. Thus, finding suitable ways to represent point cloud data structure while keeping its own internal physical property such as permutation and scale-invariant is a fundamental problem.
Therefore, we propose our work, DRINet, which serves as the basic network structure for dual-representation learning with great flexibility at feature transferring and less computation cost, especially for large-scale point clouds. DRINet mainly consists of two modules called Sparse Point-Voxel Feature Extraction and Sparse Voxel-Point Feature Extraction. By utilizing these two modules iteratively, features can be propagated between two different representations. We further propose a novel multi-scale pooling layer for pointwise locality learning to improve context information propagation.
Our network achieves state-of-the-art results for point cloud classification and segmentation tasks on several datasets while maintaining high runtime efficiency.
For large-scale outdoor scenarios, our method outperforms state-of-the-art methods with a real-time inference time of $62ms$ per frame.

\end{abstract}

\section{Introduction}

\let\thefootnote\relax\footnotetext{\textsuperscript{$\ddagger$}Part of the work was done during an internship at DEEPROUTE.AI.}
\let\thefootnote\relax\footnotetext{\textsuperscript{$*$}Equal contributions.}

\begin{figure}
    \centering
    \includegraphics[width=1\linewidth]{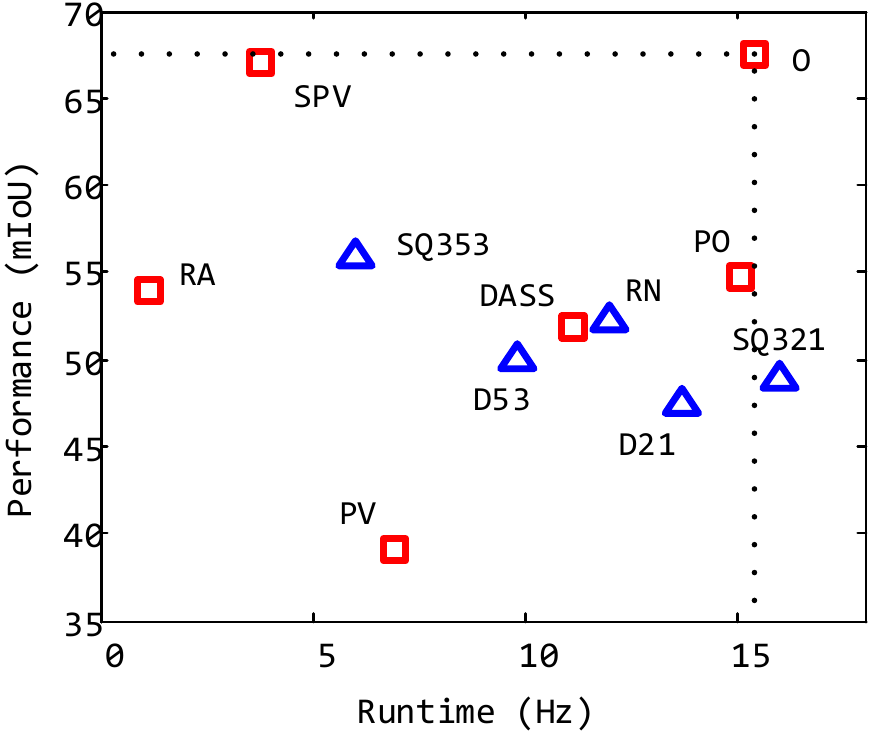}
    \caption{The mIoU performance vs. speed on the SemanticKITTI test set. Projection methods are drawn as blue triangles and other kinds of methods are drawn as red rectangles. Methods close to the right top location mean that achieving better performance within less runtime cost. Drawn methods are RA: RandLA~\cite{hu2020randla}, PV: PVCNN~\cite{liu2019point}, SPV: Sparse PVCNN~\cite{tang2020searching}, DASS~\cite{unal2021improving} PO: PolarNet~\cite{zhang2020polarnet}, D53: Darknet53~\cite{behley2019SemanticKITTI}, D21: Darknet21~\cite{behley2019SemanticKITTI}, RN: RangeNet++~\cite{milioto2019rangenet++}, SQ321: SqueezeSegV3-21~\cite{wu2018squeezeseg}, SQ353: SqueezeSegV3-53~\cite{wu2018squeezeseg}, O: our DRINet. Our DRINet outperforms all the existing methods while maintaining high runtime efficiency at $15$Hz.}
    \label{fig:P-R}
    \vspace{-5px}
    \end{figure}


Point cloud data plays a significant role in various real-world applications, from autonomous driving to augmented reality (AR). One of the critical tasks in point cloud understanding is point cloud semantic segmentation, which can facilitate self-driving cars or AR applications to interact with the physical world. For real-world applications, an accurate and real-time point cloud segmentation method is highly desirable. Therefore, in this work, we will study a new framework for high-quality point cloud segmentation in real-time.

\begin{figure}[t]
    \vspace{-10px}
    \centering
    \includegraphics[width=8.5cm]{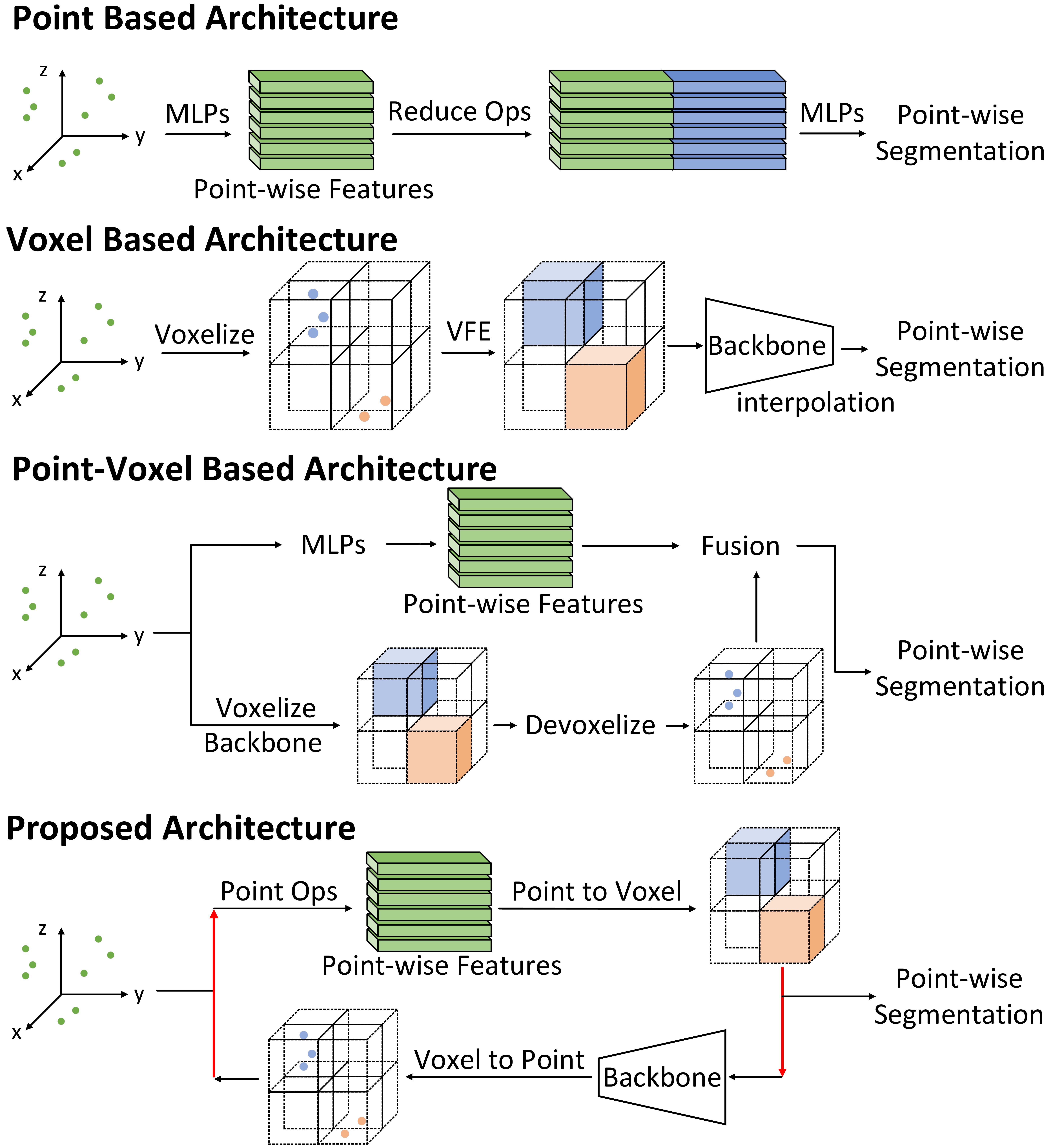}
    \caption{Three common structures of the 3D semantic segmentation task (Point Based Architecture, Voxel Based Architecture, Point-Voxel Based Architecture), and the difference compared with our proposed architecture. Notice the arrow direction of red lines that represents that dual-branches are integrated iteratively.}
    \label{fig:intro}
    \vspace{-5px}
\end{figure}
Although we have witnessed great progress in vision tasks on 2D images with convolutional neural networks (CNN), point cloud processing with deep learning still faces lots of challenges. Due to its sparsity and irregularity, it is difficult to directly apply 2D CNNs or some other popular operations in image processing for point cloud data. PointNet~\cite{qi2017pointnet} is a pioneering work that directly operates on raw point clouds. PointNet++~\cite{qiPointNetDeepHierarchical2017a} extends the PointNet by aggregating local features at different scales of neighborhoods to capture more context information and fine geometry structures. Further, VoxelNet~\cite{zhouVoxelNetEndtoEndLearning2018} firstly combines learning-based point cloud feature extraction with a standard CNN structure. However, these works cannot achieve the balance between efficiency and performance, especially in large-scale outdoor scenarios where the point number in a point cloud is large.
PointNet~\cite{qi2017pointnet} or  PointNet++~\cite{qiPointNetDeepHierarchical2017a} requires a lot of memory usage and computational cost. A key hyper-parameter in VoxelNet~\cite{zhouVoxelNetEndtoEndLearning2018} is the voxel scale: a small voxel scale brings better performance but along with less runtime efficiency. Recently, Liu et al. proposed PVCNN~\cite{liu2019point} based on a dual representation that combines the merits of 3D CNN and PointNet~\cite{qi2017pointnet}. It deeply fuses the voxelwise and pointwise features. Also, SPVNet~\cite{tang2020searching} applies a similar idea by searching architecture while replacing 3D CNN layers with a 3D sparse CNN to achieve less memory cost and better computational efficiency. Both works are aiming to obtain a better feature for scene understanding tasks by dual-representation fusion. However, there are three main drawbacks of these two works. 
First, they only apply a simple fusion strategy and ignore the feature propagation among these representations, which can be mutually complementary. Second, they ignore some physical properties such as accurate measurement information of point clouds that can bring internal scale invariance. Third, they use bilinear or trilinear gathering operations to fetch pointwise features from voxel feature maps that can be a large overhead when dealing with large-scale point clouds.

Inspired by PVCNN~\cite{liu2019point}, SPVNet~\cite{tang2020searching} and considering these aspects, we propose the DRINet that serves as a better and novel framework for dual representations point cloud segmentation learning. Our DRINet has better flexibility in converting between dual representations, with which we can learn features iteratively between point and voxel representations (shown in the Fig.~\ref{fig:intro}) by our proposed novel modules: Sparse Point-Voxel Feature Extraction (SPVFE) and Sparse Voxel-Point Feature Extraction (SVPFE). Each module takes the features of the other module as input. As such, we can preserve the fine details by pointwise features and explore more context information with large receptive fields by voxelwise features. Beyond these two modules, we explore multi-scale feature extraction and aggregation for pointwise feature learning in our SPVFE to maintain its locality for better context information. Furthermore, we replace the bilinear and trilinear gathering operations with an attentive gathering layer to reduce the computation cost of feature transformation from voxelwise features to pointwise features under the SVPFE module while maintaining the performance. 




In summary, our contributions include
\begin{itemize}
\item We propose a novel network architecture for point cloud learning that can flexibly transform representations between pointwise and voxelwise features. Both pointwise and voxelwise features can be aggregated and propagated iteratively.
\item A multi-scale pooling layer is proposed at the voxel level to efficiently extract multi-scale pointwise features to gain better context information of point clouds. 
\item We propose a novel attentive gathering layer to gain better pointwise features from voxel features at a low memory access cost.
\item To demonstrate the effectiveness of our method, extensive experiments are conducted on both indoor and outdoor datasets including ModelNet~\cite{7298801}, ShapeNet~\cite{7298801}, S3DIS~\cite{armeni2017joint}, and SemanticKITTI~\cite{behley2019SemanticKITTI}. Compared with existing methods, our DRINet achieves the state-of-the-art performance on SemanticKITTI, one of the most challenging datasets for outdoor scene parsing, while running at a real-time speed of $62ms$ per frame on an Nvidia RTX 2080 Ti GPU.
\end{itemize}

\section{Related Work}
\vspace{-5px}

Data representation is a key component in point cloud related tasks, including 3D object detection and 3D semantic segmentation. Most existing works for point cloud processing can be roughly divided into the following four categories according to their representations.

\textbf{Point based methods.} Most point based works can be viewed as extensions of PointNet~\cite{qi2017pointnet} and PointNet++~\cite{qiPointNetDeepHierarchical2017a}. They usually use the farthest sampling to sample some key points to reduce computation costs when dealing with large-scale outdoor point clouds. Then a series of variant point convolution operations like PointConv~\cite{wu2019pointconv} are applied to points within the given neighborhood to extract global and local context information based on PointNet architecture. However, there are two main drawbacks to this kind of method. First, the performance of these works somehow is limited by the procedure of farthest sampling that is proposed to reduce the memory cost and increase runtime efficiency. Thus, KPConv~\cite{thomas2019kpconv} introduces a new learnable way to generate kernel points rather than farthest sampling, with better and more robust distributions to represent their local neighborhood properties. RandLA-Net~\cite{hu2020randla} also proposes a random sampling strategy to improve the efficiency of point cloud pre-processing significantly. Secondly, most of these methods heavily rely on \textit{K-nearest neighbor} search to maintain the local relationship among points per frame which involves \textit{KD Tree} building whose worst time complexity is ${\rm O}\left( {Kn\log n} \right)$.

\textbf{Projection networks.} Currently a lot of works~\cite{chen2019suma++,cortinhal2020salsanext,milioto2019rangenet++,wu2018squeezeseg,lawin2017deep} project points to front view representations including depth image and spherical projections. With this representation whose data organization is regular and structural, a series of standard convolution layers and recent popular 2D segmentation backbone~\cite{howard2017mobilenets,tan2019efficientnet,zhao2017pspnet} can be directly applied to achieve the balance between efficiency and accuracy. For example, SqueezeSeg~\cite{wu2018squeezeseg} uses spherical projections and improves the segmentation network by their SAC module. The final results are refined by the CRF process. RangeNet++~\cite{milioto2019rangenet++} uses a similar projection method with better post-processing algorithms. However, the performance of these methods is highly related to projection resolution and the complex post-processing stage that aims to smooth and refine the prediction results with extra computation cost.

\textbf{Voxel based methods.} BEV representation with regular cartesian coordinate is the most common and popular way in voxel based method. Most works in lidar detection and segmentation~\cite{langPointPillarsFastEncoders2019,zhouVoxelNetEndtoEndLearning2018} adopt this way to form 2D/3D birdeye view image features. One of the biggest advantages of this method is that it can maintain the physical properties of point clouds and apply standard convolution layers. Recently, PolarNet~\cite{zhang2020polarnet} introduces polar representation into deep learning, presenting point cloud as a ring based structure. A ring CNN is proposed to track the special data distribution properties. Since most outdoor point clouds are obtained from scanning, this representation can reduce the effect of the non-uniform distribution phenomenon compared with normal voxelization methods. Furtherly, Cylinder3D~\cite{zhou2020cylinder3d} extends the 2D polar to 3D polar voxels. Besides, Su \textit{et al}~\cite{su2018splatnet} voxelized points into the high dimensional space lattice and apply bilateral convolutions to the occupied sectors of the lattice.

\textbf{Multiview fusion based methods.} MV3D~\cite{Chen2016Multi} is the pioneering work that explored the potential of multiview features learning in 3D object detection. With PointNet, VoxelNet~\cite{zhouVoxelNetEndtoEndLearning2018} and PVCNN~\cite{liu2019point} integrated point feature with 3D volumetric representation. 3D CNN and MLP were used for extracting coarse and fine-grained features respectively to achieve better performance and less memory cost. Besides, a lot of works ~\cite{chenFastPointRCNN2019,kuImproving3DObject2019,langPointPillarsFastEncoders2019,wangVoxelFPNMultiscaleVoxel2019} have utilized point based methods for feature extraction at each single voxel rather than handcrafted feature. They all address the importance of representation integration and fusion.

\noindent In comparison with these methods above, our proposed method belongs to multiview representation learning. While taking advantage of voxel feature learning and point feature learning, our method greatly improves the segmentation performance at very high runtime efficiency. 

\begin{figure*}[t]
    \vspace{-5px}
    \centering
    \includegraphics[width=1.0\textwidth]{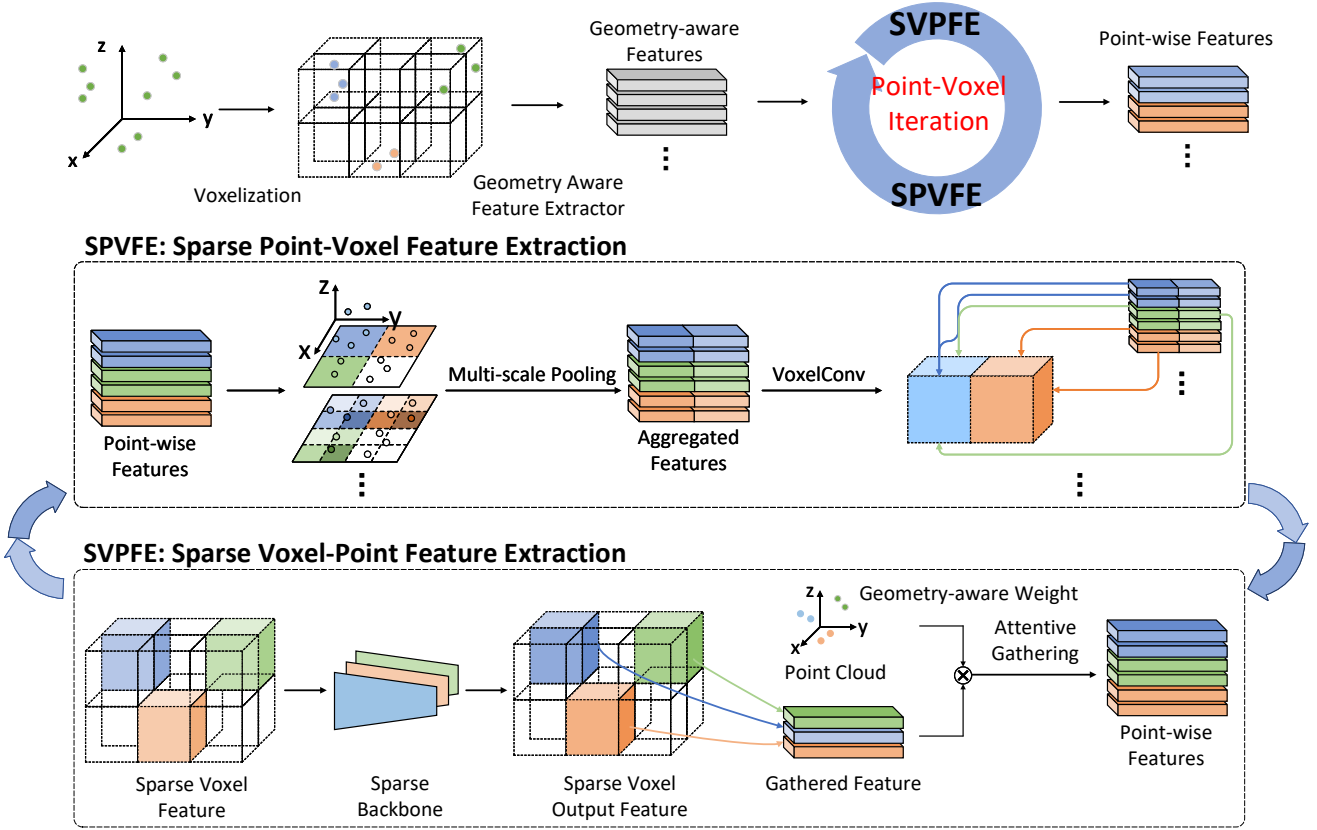}
    \caption{The first line is the whole network structure of DRINet. It includes two main modules, 1) Geometry-aware Feature Extraction, and 2) the Point and Voxel Branch. The second line describes the process of Point and Voxel Branch, consisting of Sparse Point-Voxel Feature Extraction~\textit{(SPVFE)} and Sparse Voxel-Point Feature Extraction~\textit{(SVPFE)}. a)~\textit{SVPFE} generates pointwise features with an attentive gathering layer from voxelwise features. b)~\textit{SPVFE} generates voxelwise features at target scale with a multi-scale pooling layer from pointwise features.}
    \label{fig:network}
    \vspace{-5px}
\end{figure*}

\section{Method}
\vspace{-5px}

In this section, we introduce our DRINet that integrates the merits of point and voxel representations to improve point cloud segmentation performance while maintaining high computational efficiency. 
The overall network, as shown in Fig.~\ref{fig:network}, consists of four parts: 1) \textbf{Geometry-aware Feature Extraction} 2) \textbf{Sparse Voxel-Point  Feature  Extraction} 
3) \textbf{Sparse Point-Voxel  Feature  Extraction} and 4) \textbf{Iterative Dual-Representation Learning}. The sparse point-voxel feature extraction layer takes pointwise features as input and outputs voxelwise features to form sparse voxel feature maps with more hierarchical information. 
Then the sparse voxel-point feature extraction layer takes voxelwise features as input to generate high-quality pointwise features. The two blocks can iteratively perform the conversion between different representations, namely \textbf{iterative dual-representation learning}.

\vspace{-5px}
\subsection{GAFE: Geometry-aware Feature Extraction}
\vspace{-5px}

\label{sec:gafe}
\textbf{Data Representation.}
A point cloud can be represented by an unordered point set $\left\{ {{{{p}}_1},{{{p}}_2}, \ldots ,{{{p}}_N}} \right\}$ with ${{{p}}_i} \in {\mathbb{R}^d}$ that includes the point coordinate $c_i = (x_i, y_i, z_i)$ and associated point features such as intensity.

\textbf{Voxelization.} 
We introduce the voxelization process to construct a mapping relationship between the two representations. Define that the point cloud is discretized into numerous voxels with resolution of ${L} \times {W} \times {H}$ and $N_V$ non-empty voxel numbers. Given a point ${{{p}}_i}$, 
we compute its voxel index ${{v}_i}$ under the grid scale $s$: 
\begin{equation}
    \setlength{\abovedisplayskip}{1pt}
    \setlength{\belowdisplayskip}{1pt}
        {{{v}}_i^{s}} = 	\left(\lfloor{x_i/s}\rfloor, \lfloor{y_i/s}\rfloor, \lfloor{z_i/s}\rfloor \right),
\label{equ:voxel_p}
\end{equation}
where $\lfloor{\cdot}\rfloor$ is the floor function and $s$ refers to the size of each voxel along $xyz$ directions. 

\textbf{Scatter} ${{\Phi}^s_{\mathcal{P} \to \mathcal{V}}}$ and \textbf{Gather} ${{\Phi}^s_{\mathcal{V} \to \mathcal{P}}}$. Now a mapping system for coordinate spaces between point $p$ and voxel $v$ has been built for indexing.
We define two flexible operations $\bf{Scatter}$ ${{\Phi}^s_{\mathcal{P} \to \mathcal{V}}}$ and $\bf{Gather}$ ${{\Phi}^s_{\mathcal{V} \to \mathcal{P}}}$ to transform between voxelwise features $\mathcal{V}^s$ and pointwise features $\mathcal{P}$ under the voxel scale $s$, where $\mathcal{V}^s \in {R^{N_V \times C}}$, $\mathcal{P} \in {R^{N \times C}}$, and $C$ is the number of channels. For the \textbf{Scatter} operation, under the voxel scale $s$, the voxel feature at the voxel $\zeta$ is obtained by a boardcast operation $\Psi$ on all points inside this voxel:
\begin{eqnarray}
    \mathcal{V}^s=\{{\mathcal{V}_{\zeta}^s} \}={{\Phi}^s_{\mathcal{P} \to \mathcal{V}}}\left( {\mathcal{P}} \right),\quad 
    \mathcal{V}_{\zeta}^s= {\Psi}\left( {\left\{ {{ {\mathcal{P}}}{_i}|v_i^{s} = \zeta} \right\}} \right),
\label{equ:scatter}
\end{eqnarray}
where ${\Psi}$ can be defined as the mean or max operation. In brief, ${\Phi}^s_{\mathcal{P} \to \mathcal{V}}$ can broadcast apply the same operation for all the input points within the same voxel.
Meanwhile, we define an inverse operation \textbf{Gather}. The $i$-th pointwise feature with voxel $\zeta$ is gathered from the voxelwise features by a boardcast identity mapping operation (i.e., copying):
\begin{eqnarray}
    \mathcal{P}=\{\mathcal{P}_i\}={{\Phi}^s_{\mathcal{V} \to \mathcal{P}}}\left( {\mathcal{V}^s}
     \right),\quad \mathcal{P}_i=\mathcal{V}_{\zeta}^s.
\label{equ:gather}
\end{eqnarray}


\label{sec:raw_pfe}
\textbf{Geometry-aware Feature Extraction.} Inspired by works~\cite{langPointPillarsFastEncoders2019, ye2020hvnet}, we focus on fully utilizing the original point cloud geometric properties. The raw geometry-aware feature $g^s_i$ for point $p_i$ under a given grid size $s$ is represented as 
\begin{equation}
    {g}^s_i = {( {{{c}_i} - {{\sum\nolimits_{{c_j} \in \mathcal{N}_i^s} {{c_j}} } \mathord{\left/
 {\vphantom {{\sum\nolimits_{{V_i}} {{c_j}} } {{N_{V_i}}}}} \right.
 \kern-\nulldelimiterspace} {|{\mathcal{N}_i^s}|}}} ) \oplus {{p}_i}} \oplus \left( {{{c}_i} - s\times{v}^s_i} \right),
\label{equ:raw_feature}
\end{equation}
where $\oplus$ represents tensor concatenation. The neighbor collection $\mathcal{N}_i^s$, referred to the point coordinates set of points that lies in the same voxel as ${{p}}_i$, is denoted as $\mathcal{N}_i^s = \left\{ {c_j |v_j^s = v_i^s} \right\}$. 
Let $G^s=\{g_i^s\}$ and then the final multi-scale feature $G$ is
\begin{align}
    {G}{\text{ }} = \sum\limits_{s \in S} {\text{MLP}({G^s}) \oplus {\Phi^s_{\mathcal{V} \to \mathcal{P}}}\left( {{\Phi^s_{\mathcal{P} \to \mathcal{V}}}\left( {\text{MLP}({G^s})} \right)} \right)},
\label{equ:feature_at_s}
\end{align}
where $\text{MLP}$ represents a multilayer perceptron, $S$ is the scale list. By simply fusing multi-scale pointwise features, we obtain the hybrid geometry-aware pointwise features $G$, which serve as the initial pointwise features $F$ for SPVFE and SVPFE.

\vspace{-5px}
\subsection{SPVFE: Sparse Point-Voxel Feature Extraction}
\vspace{-5px}

\label{sec:voxel_block}
In this part, we propose our novel Sparse Point-Voxel Feature Extraction module (SPVFE). By taking pointwise features, SPVFE provides a novel way \textbf{Multi-scale Pooling} for multi-scale pointwise features learning in point clouds with better efficiency that serves as better context information. Finally, it transforms the pointwise features into voxelwise features with our proposed \textbf{VoxelConv}. 

\textbf{Multi-scale Pooling Layer.}
To obtain better pointwise features, we propose a novel layer to explore more context information with great efficiency. Inspired by PointNet++~\cite{qiPointNetDeepHierarchical2017a}, PSPNet~\cite{zhao2017pspnet}, HVNet~\cite{ye2020hvnet}, and Deeplab~\cite{chen2017deeplab}, we notice that multi-scale information is important in classification and segmentation tasks. To some extent, multi-scale information can aggregate more local context information at different scales with different receptive fields. Indeed, pyramid pooling and dilated convolution are straightforward ways in image-related tasks. 
However, applying these methods for sparse voxel feature maps will decrease efficiency and deteriorate the feature since many empty voxel features will be involved. Therefore, we propose a novel multi-scale pooling layer that utilizes simple MLP layers for point clouds followed by set abstraction at different scales by the scattering operation in Eq.~\ref{equ:scatter}. 
As shown in Alg.~\ref{algor:multi-scale-pooling}, for each scale $s$ in the given scale list $S$, only points inside the same voxel will contribute to the output features. With the aggregation of pointwise features at different scales, there will be a stronger representation of multi-scale properties in point clouds. Compared with PointNet++~\cite{qiPointNetDeepHierarchical2017a}, our method does not rely on KNN that consists of a complex pre-processing procedure for building the KD Tree. 

\begin{algorithm}[htbp]
  \caption{Multi-scale Pooling Algorithm}
  \label{algor:multi-scale-pooling}
  \begin{algorithmic}[1]
    \Require
    Pointwise features $F$ and predefined scales $S$
    \State $L$ = []
    \For{each $s\in S$}
      \State $\mathcal{V}^{s} = {{\Phi}^s_{\mathcal{P} \to \mathcal{V}}}\left( {F} \right)$
      \State $F^{s} = \text{MLP}(\text{Concat}([F, {{\Phi}^s_{\mathcal{V} \to \mathcal{P}}}\left( \mathcal{V}^s \right)]))$
      \State $L.\text{append}(F^{s})$
    \EndFor
    \State\Return $\text{Concat}(L)$
  \end{algorithmic}
\end{algorithm}

\textbf{VoxelConv Operation.} With pointwise features from the Multi-scale Pooling Layer, we propose the VoxelConv operation to form the next-stage voxel feature map.
Similar to graph convolutional networks~\cite{niepert2016learning}, we deal with points in the discrete voxel space with local aggregation. VoxelConv in the voxel space with pointwise features to generate the voxelwise features can be defined as
\begin{equation}
     {{\Phi}^s_{\mathcal{P} \to \mathcal{V}}}\left( {W F} \right),  
\end{equation}
where $W$ is a weight matrix. VoxelConv is applied to all the points within the same voxel, where we use fully connected layers with learnable weights $W$ to each pointwise feature. Then the aggregation function in Eq.~\ref{equ:scatter} is used to calculate the local response. The whole process is similar to graph convolutions ~\cite{niepert2016learning} with learnable weights and maintains locality. {VoxelConv} generates sparse voxel features and skips empty voxel grids, saving a lot of computation costs. Finally, we can construct the feature map by assigning the voxel features to their corresponding locations.


\vspace{-5px}
\subsection{SVPFE: Sparse Voxel-Point Feature Extraction}
\vspace{-5px}

\label{sec:point_block}
Taking 3D voxel features as input, there are several mature ways to exploit the local spatial correlation of the input in the 3D space. PVCNN~\cite{liu2019point} utilizes 3D CNN layers. However, it neglects the large-scale outdoor scenarios that contain hundreds of thousands of points per frame. 
As such, we propose a dual module, namely SVPFE module against SPVFE, with which the voxelwise features from SPVFE modules are fed into a 3D Sparse Voxel Learning block and then rollback to pointwise by an attentive gathering layer.

\textbf{3D Sparse Voxel Learning.} 
Most previous works compress the Z-axis information for fast feature extraction without large GPU memory usage but with an aggressive downsampling strategy for the sake of higher efficiency, leading to ineffectiveness in capturing small instances.
Other works deploy 3D features but suffer the computation inefficiency.
Inspired by sparse convolution~\cite{Graham20173D, yan2018second}, we adopt 3D sparse convolution as backbone for voxel features extraction due to its high efficiency and ability to capture compact voxel features.
We utilize a series of ResNet Bottleneck~\cite{he2016deep} by replacing 2D convolution with 3D sparse convolution. We name it Sparse Bottleneck.

\textbf{Attentive Gathering Strategy.} 
After Sparse Bottleneck, we need to map the voxelwise features to pointwise features. A nearest gathering operation that retrieves pointwise features from the voxelwise features is applied according to Eq.~\ref{equ:gather}.
Since the points in the same voxel share the same voxel feature, the nearest gathering will lead to inferior feature representation capability. 
Especially when the voxel scale increases, meaning that each voxel will contain more points, this phenomenon will become more severe. 
Previous works~\cite{liu2019point, tang2020searching} adopt bilinear or trilinear gathering operations when retrieving the pointwise features. 
However, the memory access cost for a large number of points cannot be ignored since it could not guarantee the memory coalescing that allows an optimal usage of the global memory bandwidth. 
There will be a great overload for the whole model once more gathering operations are introduced.

\begin{figure}
    \vspace{-10px}
\centering
\hspace*{-3mm}
\includegraphics[width=8.8cm]{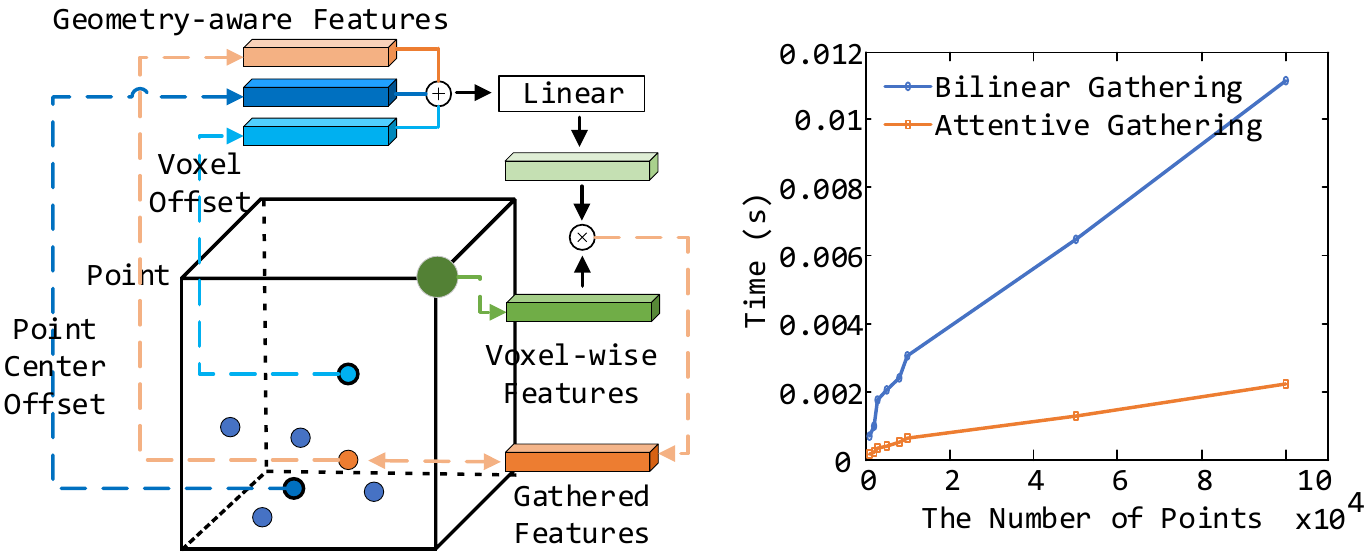}
\caption{The left figure illustrates our attentive gathering. The right one shows the computational cost for the bilinear gathering and our attentive gathering.}
\label{fig:gather}
\vspace{-10px}
\end{figure}

Thus, we propose a novel and effective approach with learnable parameters to increase the uniqueness and representation capability while maintaining the voxelwise features, as shown in Fig.~\ref{fig:gather}. The traditional bilinear gathering can be expressed as weighted sum of neighborhood features according to distance.
As a contrast, our method can be derived as follows:
\begin{gather}
    F_{att} = W' G, \\
    F_{out} = F \odot F_{att}.
\end{gather}
where $G$ is the hybrid geometry-aware features from Sec.~\ref{sec:raw_pfe}, and $W'$ is a weight matrix. $F_{att}$ is geometry-aware weights. Then output features are obtained by elementwise multiplication of nearest gathering features $F$ and above geometry-aware weights.
Compared with bilinear gathering, our attentive gathering layer contains statistics, including mean and voxel information as geometric prior information with learnable parameters, which can be viewed as an attention mechanism.

\vspace{-5px}
\subsection{Iterative Dual-Representation Learning}
\vspace{-5px}

\begin{algorithm}[t!]
    \caption{Dual-representation Learning Algorithm}
    \label{algor:dual_rep}
    \begin{algorithmic}[1]
      \Require
      Point cloud $P$, \#Iteration $N_I$, Scale list $S$
      \Ensure
      Pointwise semantic prediction $O_p$
      \State $F$ = []
      \State $G \leftarrow \text{GAFE}(P, S)$
      \State $F_p \leftarrow G$
      \For{$\text{iter}=1$  to  $N_I$}
        \State $F_v \leftarrow \text{SPVFE}(F_p)$
        \State $F_p \leftarrow \text{SVPFE}(F_v)$
        \State $F.\text{append}(F_p)$
      \EndFor
      \State $O_p \leftarrow \text{SoftMax(}\text{MLP}(\text{Concat}(F)))$
    \end{algorithmic}
\end{algorithm}

Most previous works only utilize single form of features such as pointwise~\cite{hu2020randla}, voxelwise~\cite{zhang2020polarnet}, etc., or extract multi-representation in parallel. On the contrary, we propose a novel dual-representation learning algorithm based on an iterative process shown in Alg.~\ref{algor:dual_rep}. 
The hybrid geometry-aware features $G$ from Sec.~\ref{sec:gafe} are fed into the iteration process, composed with SVPFE module and SPVFE module. The SVPFE module contains sparse convolution layers that can learn sufficient intra-voxel features, while the SPVFE module is capable of capturing inner-voxel features with the geometry constraint for points inside the same voxel with multi-scale pooling for better locality and context extraction. Both SPVFE and SVPFE modules utilize the output of the other module. As such, pointwise features and voxelwise features propagate mutually and iteratively, forming a natural foundation for fusing context information across multiple representations. 

\begin{figure*}[t]
    \vspace{-10px}
    \centering
    \includegraphics[width=1.0\textwidth]{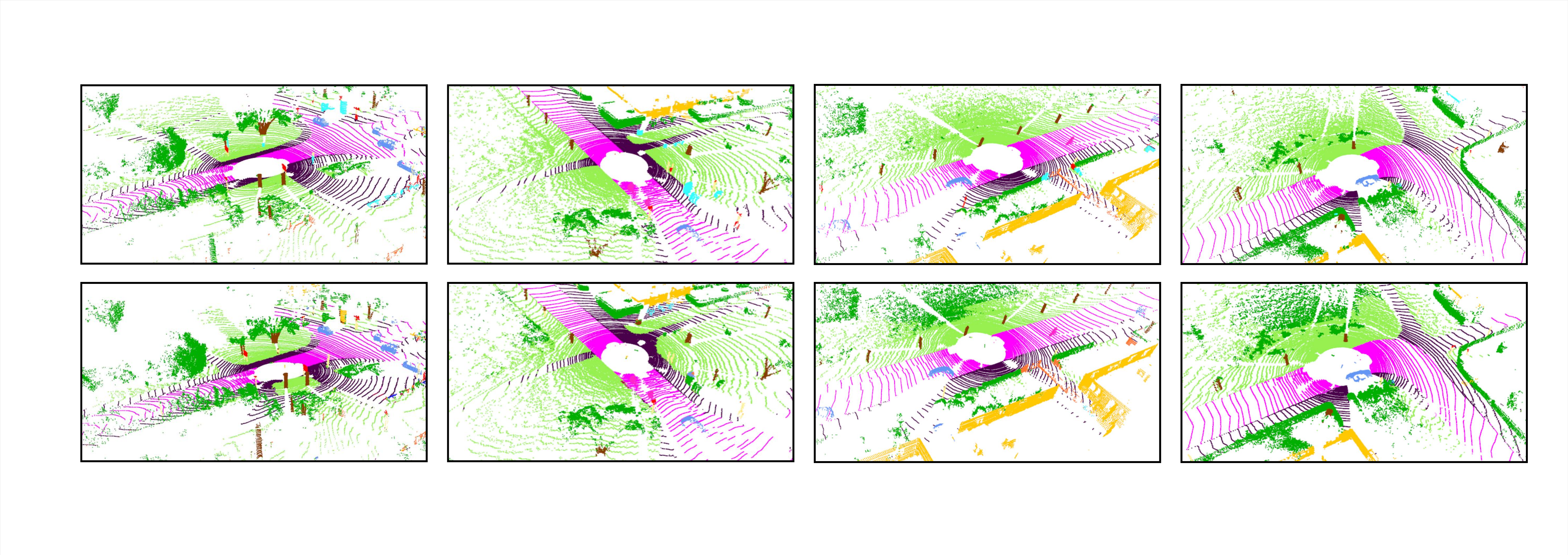}
    \caption{The results on SemanticKITTI. The top row is the ground truth, and the bottom row is the Predictions by DRINet.}
    \label{fig:semantic_results}
    \vspace{-5px}
\end{figure*}

\begin{table*}[!t]
    \small
    \centering
    \resizebox{\textwidth}{!}{
\setlength{\tabcolsep}{1.3mm}{      \def\arraystretch{1.1}
    \begin{tabular}{@{}l c c c c c c c c c c c c c c c c c c c c c@{}}
    \hline
    Methods & \rotatebox{90}{road} & \rotatebox{90}{sidewalk} & \rotatebox{90}{parking} & \rotatebox{90}{other ground} & \rotatebox{90}{building} & \rotatebox{90}{car} & \rotatebox{90}{truck} & \rotatebox{90}{bicycle} & \rotatebox{90}{motorcycle} & \rotatebox{90}{other vehicle \,} & \rotatebox{90}{vegetation} & \rotatebox{90}{trunk} & \rotatebox{90}{terrain} & \rotatebox{90}{person} & \rotatebox{90}{bicyclist} & \rotatebox{90}{motorcyclist} & \rotatebox{90}{fence} & \rotatebox{90}{pole} & \rotatebox{90}{traffic sign} & \rotatebox{90}{mIoU} & \rotatebox{90}{speed (ms)}
    \\
    \hline
    PointNet~\cite{qi2017pointnet} & 61.6 & 35.7 &	15.8 & 1.4 & 41.4 &	46.3 & 0.1 & 1.3 & 0.3 & 0.8 & 31.0 & 4.6 & 17.6 & 0.2 & 0.2 & 0.0 & 12.9 & 2.4 & 3.7 & 14.6 & 500
    \\
    \hline
    PointNet++~\cite{qiPointNetDeepHierarchical2017a} & 72.0 & 41.8 &	18.7 & 5.6 & 62.3 &	53.7 & 0.9 & 1.9 & 0.2 & 0.2 & 46.5 & 13.8 & 30.0 & 0.9 & 1.0 & 0.0 & 16.9 & 6.0 & 8.9 & 20.1 & 5900
    \\
    \hline
    KPConv~\cite{thomas2019kpconv} & 88.8 & 72.7 &	61.3 & \textbf{31.6} & 90.5 &	96.0 & 33.4 & 30.2 & 42.5 & 44.3 & 84.8 &	69.2 & 69.1 & 61.5 & 61.6 & 11.8 & 64.2 & 56.4 & 47.4 & 58.8 & -
    \\
    \hline
    SqueezeSegV3~\cite{xu2020squeezesegv3} & 91.7 & 74.8 & 63.4 & 26.4 & 89.0 & 92.5 & 29.6 & 38.7 & 36.5 & 33.0 & 82.0 & 58.7 & 65.4 & 45.6 & 46.2 & 20.1 & 59.4 & 49.6 & 58.9  & 55.9 & 238
    \\
    \hline
    TangentConv~\cite{tatarchenko2018tangent} &83.9 & 63.9 &33.4 &15.4& 83.4& 90.8& 15.2& 2.7& 16.5& 12.1& 79.5& 49.3& 58.1& 23.0 & 28.4 & 8.1& 49.0& 35.8& 28.5 & 35.9 & 3000
    \\
    \hline
    SPVNet~\cite{tang2020searching} & 90.2 & \textbf{75.4} & \textbf{67.6} & 21.8 & \textbf{91.6} & \textbf{97.2} & 56.6 & 50.6 & 50.4 & \textbf{58.0} & \textbf{86.1} & \textbf{73.4} & \textbf{71.0} & 67.4 & 67.1 & 50.3 & 66.9 & \textbf{64.3} & \textbf{67.3} & 67.0 & 259
    \\
    \hline
    PolarNet~\cite{zhang2020polarnet} & 90.8 & 74.4 & 61.7 & 21.7 & 90.0 & 93.8 & 22.9 & 40.3 & 30.1 & 28.5 & 84.0 & 65.5 & 67.8 & 43.2 & 40.2 & 5.6 & 61.3 & 51.8 & 57.5 & 54.3 & \textbf{62}
    \\
    \hline
    RandLA~\cite{hu2020randla} & 90.7 & 73.7 & 60.2 & 20.4 & 86.9 & 94.2 & 40.1 & 26.0 & 25.8 & 38.9 & 81.4 & 66.8 & 49.2 & 49.2 & 48.2 & 7.2 & 56.3 & 47.7 & 38.1 & 53.9 & 880
    \\
    \hline
    RangeNet++~\cite{milioto2019rangenet++} & 91.8 & 75.2 & 65.0 & 27.8 & 87.4 & 91.4 & 25.7 & 25.7 & 34.4 & 23.0 & 80.5 & 55.1 & 64.6 & 38.3 & 38.8 & 4.8 & 58.6 & 47.9 & 55.9 & 52.2 & 83.3
    \\
    \hline
    DASS~\cite{unal2021improving} & \textbf{92.8} & 71.0 & 31.7 & 0.0 & 82.1 & 91.4 & \textbf{66.7} & 25.8 & 31.0 & 43.8 & 83.5 & 56.6 & 69.6 & 47.7 & 70.8 & 0.0 & 39.1 & 45.5 & 35.1 & 51.8 & 90
    \\
    \hline
    \textbf{DRINet}\textbf{(ours)} & 90.7 & 75.2 & 65.0 & 26.2 & 91.5 & 96.9 & 43.3 & \textbf{57.0} & \textbf{56.0} & 54.5 & 85.2 & 72.6 & 68.8 & \textbf{69.4} & \textbf{75.1} & \textbf{58.9} & \textbf{67.3} & 63.5 & 66.0 & \textbf{67.5} & \textbf{62}
    \\
    \hline
    \end{tabular}}}
    \caption{The per-class mIoU results on the SemanticKITTI test set.}
    \label{semantic_kitti}
\end{table*}

\section{Experiments}
\vspace{-5px}
We conduct extensive experiments for the proposed DRINet on both indoor and outdoor tasks, including classification and segmentation, to show the effectiveness and generalization ability of our proposed method.

\vspace{-5px}
\subsection{Outdoor Scene Segmentation}
\vspace{-5px}

\textbf{Dataset.} We use the SemanticKITTI~\cite{behley2019SemanticKITTI} dataset to verify the effectiveness of our network for large-scale outdoor scenarios. SemanticKITTI has a total of 43551 scans with imbalanced point level annotations for 20 categories. It contains 22 sequences which involve the most common scenes for autonomous driving.
Besides, another challenging part of this dataset is that each scan contains more than 100K points on average, posing great pressure on lightweight model design. Following the official settings, we use the sequences from 00 to 10 except 08 as the training split, sequence 08 as validation split, and the sequences from 11 to 21 as the test split.



\textbf{Experiment Details.} Although the maximum distance for point clouds in SemanticKITTI can be more than 80m with a non-uniform density distribution,
there are few points when sensing range beyond 55m. Based on this observation, we set voxelization scale ranging from minimum $[-48, -48, -3]$ to maximum $[48, 48, 1.8]$ 
for x, y, z respectively, with which we can include nearly $99$\% 
points with only $1$\% mIoU lost.
For the points outside the ranges, we mask them to unknown types.

\textbf{Network Details.} Following the above principle, we design DRINet with three SPVFE and SVPFE blocks with the voxel scales at $[0.4m, 0.8m, 1.6m]$ respectively and Multi-scale Pooling Layer at $[0.4m, 0.8m, 1.6m, 3.2m]$. We also do some experiments varying the number of SPVFE and SVPFE blocks. In the loss design, we adopt the Lovasz loss~\cite{berman2018lovasz} to alleviate the great imbalance distribution among different categories. During training, global rotation and random flip are applied for data augmentation. We train DRINet for 40 epochs with the Adam~\cite{kingma2014adam} optimizer with batch size of 4, the initial learning rate is $2{e^{ - 4}}$ with weight decay $1{e^{ - 4}}$. Besides, the learning rate decays with a ratio of $0.8$ every 5 epochs.

\textbf{Experimental Results.} Detailed per-class quantitative results of DRINet and other state-of-the-art methods are shown in Tab.~\ref{semantic_kitti}. DRINet achieves state-of-the-art performance among these methods in the mean IoU score. In some small classes, such as bicycle, person and so on, the DRINet shows a far bigger improvement. Moreover, we maintain a real-time inference time with the highest performance-time ratio, shown in Fig.\ref{fig:P-R}. We also provide some qualitative visual results on SemanticKITTI test set, as shown in Fig.~\ref{fig:semantic_results}.

\vspace{-5px}
\subsection{Ablation Study}
\vspace{-5px}

To analyze the effectiveness of different components in DRINet, we conduct the following ablation studies on the SemanticKITTI validation set.

\textbf{Geometry-aware Feature Extractor.} We firstly analyze our model with only geometry-aware feature extractor (GAFE), which is shown in the first line in Fig.\ref{fig:network} with the iteration part removed. Compared with PointNet~\cite{qi2017pointnet} and PointNet++~\cite{qiPointNetDeepHierarchical2017a}, our GAFE is a stronger baseline with 22.8\% mIoU on validation set which means our GAFE has better representations for the physical properties of the original data, as shown in Tab.~\ref{ablation_study_raw_extractor}.

\begin{table}[htbp]
    \small
    \centering
    \def\arraystretch{1.1}
    \begin{tabular}{c|c|c|c}
    \hline
     & PointNet~\cite{qi2017pointnet} & PointNet++~\cite{qiPointNetDeepHierarchical2017a} & GAFE
    \\
    \hline
    mIoU(\%) & 15.3 & 18.1 & \textbf{22.8}
    \\
    \hline
    Latency (s) & 0.5 & 5.9 & \textbf{0.022}
    \\
    \hline
    \end{tabular}
\vspace{1px}
    \caption{Comparison with different feature extractors (GAFE).}
    \label{ablation_study_raw_extractor}
    \vspace{-10px}
\end{table}

\textbf{Representation Analysis.} The core component of our DRINet lies in dual-representation. We remove either representation to verify how the block influences the final results. As shown in Tab.~\ref{ablation_study_representation}, we set value of SPVFE number $N_{spv}$ or SVPFE number $N_{svp}$ to zero respectively to control the representation involved. 
With SPVFE and SVPFE off, there will be about 18.7\% and 3.9\% drop respectively.
It illustrates that network with only voxelwise features performs better than that with only pointwise features. 
Moreover, by fusing dual representations, we can obtain better features.

\begin{table}[htbp]
    \small
    \centering
    \def\arraystretch{1.1}
    \begin{tabular}{c|c|c}
    \hline
    SPVFE $N_{spv}$ & SVPFE $N_{svp}$ & mIoU(\%)
    \\
    \hline
    3 & 0 & 48.6 
    \\
    0 & 3 & 63.4 
    \\
    3 & 3 & 67.3 
    \\
    \hline
    \end{tabular}
\vspace{1px}
    \caption{Ablation study of block numbers on SemanticKITTI.}
    \label{ablation_study_representation}
    \vspace{-10px}
\end{table}

\begin{figure}[htbp]
\vspace{-5px}
    \centering
    \includegraphics[width=8.5cm]{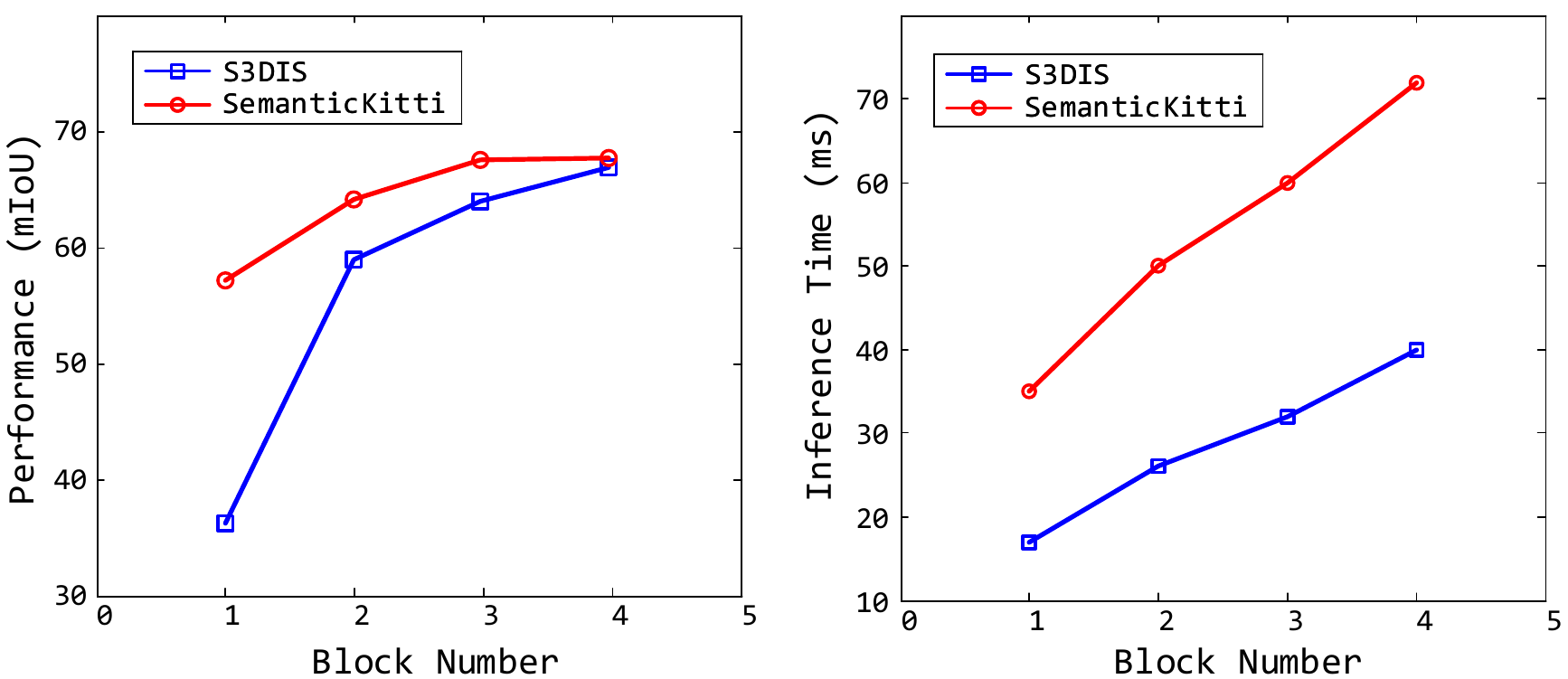}
    \caption{Performance and inference time vary with block number. We set both $N_{spv}$ and $N_{svp}$ from 1 to 4. The left / right one is Performance / Inference time vs. Block number respectively.}
    \label{fig:performance_block}
    \vspace{-10px}
    \end{figure}

\textbf{Block Number Analysis.} We note that block numbers $N_{spv}$ and $N_{svp}$ are crucial parameters for our network structure. 
From the left of Fig.~\ref{fig:performance_block}, we can see the mIoU increasing in SemanticKITTI dataset (from 58.2\% to 67.6\%) and in S3DIS dataset (from 36.3\% to 66.7\%) as block number increases. Nevertheless, this comes with more computation cost and larger memory usage. As shown in the right of Fig.~\ref{fig:performance_block}, inference time increases from 62$ms$ to 72$ms$ when add $N_{spv}$ and $N_{svp}$ from 3 to 4, however the mIoU only improved from 67.3 to 67.6. According to the balance between performance and efficiency, we finally adopt $N_{spv}=3$ and $N_{svp}=3$ in following experiments.


\textbf{Multi-scale Pooling Layers.}
Multi-scale (MS) Pooling Layer tends to aggregate more levels of context information in a wider neighborhood and different receptive fields. By removing this unit, the pointwise features will only go through several MLP Layers. Shown in Tab.~\ref{ablation_study_}, the MS Pooling Layer improves the mIoU from 65.4\% to 67.3\%.

\begin{table}[htbp]
    \small
    \centering
    \vspace{-1px}
    \def\arraystretch{1.1}
    \begin{tabular}{c|c|c|c}
    \hline
    Baseline & Attentive Gathering & MS Pooling & mIoU(\%)
    \\
    \hline
    \checkmark & $\times$ &$\times$ & 64.6
    \\
    \checkmark & \checkmark &$\times$ & 65.4
    \\
    \checkmark & \checkmark&\checkmark & 67.3
    \\
    \hline
    \end{tabular}
    \vspace{1px}
    \caption{Ablation study on SemanticKITTI.}
    \label{ablation_study_}
    \vspace{-10px}
\end{table}

\textbf{Attentive gathering.} Without attentive gathering, we only use nearest gathering due to the consideration of computation cost, and the comparison between nearest gathering and bilinear gathering is shown in Fig.~\ref{fig:gather}. With precomputed attentive weights within the voxel, the performance is improved with 0.8\% in DRINet, showing that attentive gathering can prevent the feature degradation problem.

\vspace{-5px}
\subsection{Indoor Tasks}
\vspace{-5px}

\textbf{ModelNet40 classification.} ModelNet40~\cite{7298801}, one of the most popular 3D classification datasets, contains 12,311 meshed CAD models from 40 categories. We use normals as extra features and sample 1024 points as input. Following the standard processing procedure, the input points are first normalized to unit range. As a result, the scale of multi-scale pooling is set to $[0.2m, 0.4m, 0.6m, 0.8m]$. During training, data augmentation including rotation, scaling, flipping and perturbation is applied. The DRINet is built with two SVPFE and SPVFE blocks with the voxel scales at $[0.2m, 0.2m]$. We gather all pointwise features from point blocks and apply max operation same as PointNet~\cite{qi2017pointnet}, then add one fully connected layer to generate output scores. In Tab.~\ref{model40_results}, our DRINet directly improves the performance in the 3D classification task. We observe that DRINet with dual-representations performs better than the previous state-of-the-art method with single representation. 

\begin{table}[!t]
\centering
\small
\def\arraystretch{1.1}
\begin{tabular}{c|c|c|c}
\hline
Method & Input & Size & Acc(\%)
\\
\hline
MVCNN~\cite{su15mvcnn} & \textit{I} & 3 $\times$ 1024 & 90.1
\\
\hline
PointNet++~\cite{qiPointNetDeepHierarchical2017a} & \textit{P} & 6 $\times$ 1024 & 91.9
\\
\hline
PointCNN~\cite{li2018pointcnn} & \textit{P} & 6 $\times$ 1024 & 92.2
\\
\hline
LP-3DCNN~\cite{kumawat2019lp} & \textit{V} & 6 $\times$ 1024 & 92.1
\\
\hline
LDGCNN~\cite{zhang2019linked} & \textit{G} & 6 $\times$ 1024 & 92.9
\\
\hline
KPConv~\cite{thomas2019kpconv} \textit{regid} & \textit{P} & 3 $\times$ 1024 & 92.9
\\
KPConv~\cite{thomas2019kpconv} \textit{deform} & \textit{P} & 3 $\times$ 1024 & 92.7
\\
\hline
CloserLook3D~\cite{liu2020closer} & \textit{P} & 3 $\times$ 1024 & 92.9
\\
\specialrule{1pt}{0pt}{0pt}
\textbf{DRINet}\textbf{(ours)} & \textit{P} + \textit{V} & 6 $\times$ 1024 & \textbf{93.0}
\\
\hline
\end{tabular}
\vspace{1px}
\caption{Results on ModelNet40~\cite{7298801},\textit{P}, \textit{I}, \textit{G}, \textit{V} mean point, image, graph and volumetric respectively}
\label{model40_results}
\end{table}

\begin{table}[!t]
    \small
    \centering
    \begin{tabular}{c|c|c|c}
    \hline
    Method & Input & mIoU(\%) & Latency(\bf{ms})
    \\
    \hline
    KPConv~\cite{thomas2019kpconv} \textit{regid} & \textit{P} & 86.2 & -
    \\
    KPConv~\cite{thomas2019kpconv} \textit{deform} & \textit{P} & \textbf{86.4} & -
    \\
    \hline
    PointNet++~\cite{qiPointNetDeepHierarchical2017a} & \textit{P} & 85.1 & 77.9
    \\
    \hline
    PointCNN~\cite{li2018pointcnn} & \textit{P} & 86.13 & -
    \\
    \hline
    SO-Net~\cite{li2018so} & \textit{P} & 84.9 & -
    \\
    \hline
    PVCNN~\cite{liu2019point} & \textit{P} + \textit{V} & 86.2 & 50.7
    \\
    \hline
    \textbf{DRINet}\textbf{(ours)} & \textit{P} + \textit{V} & \textbf{86.4} & 30
    \\
    \hline
    \end{tabular}
\vspace{1px}
    \caption{Results on ShapeNet}
    \label{shapnet_results}
    \vspace{-5px}
\end{table}

\textbf{ShapeNet Parts segmentation.} We also evaluate our method on the segmentation task of
ShapeNet Parts dataset~\cite{10.1145/2980179.2980238} which is a collection of 16681 point clouds of 16 categories. Since ShapeNet Parts is also generated from CAD model, we apply the same settings for data pre-processing and augmentation as ModelNet40. For this task, we set block number to 4 to increase the overall network complexity, which can boost the performance. From Tab.~\ref{shapnet_results}, our network achieves state-of-the-art according to mIoU criteria with high runtime efficiency (30ms). 

\textbf{S3DIS.} The S3DIS dataset~\cite{armeni2017joint} consists of 271 rooms belonging to 6 large-scale indoor areas with 13 classes. Following the previous works, we use Area 5 as the test set and the rest as the training set. For a fair comparison, we use the same data processing and evaluation protocol as these works~\cite{li2018pointcnn, liu2019point}. We also use four blocks of SPVFE and SVPFE modules with the voxel scales at $[0.2m, 0.2m, 0.2m, 0.2m]$. 

As shown in Tab.~\ref{S3DIS_results}, our DRINet achieves state-of-the-art compared with single representation methods. For the dual-representation methods PVCNN~\cite{liu2019point}, we even outperform its largest model with less computation cost.

\begin{table}[!t]
    \small
    \centering
    \begin{tabular}{c|c|c}
    \hline
    Methods & Input & mIoU(\%)
    \\
    \hline
    PointNet~\cite{qi2017pointnet} & \textit{P}  & 41.09
    \\
    \hline
    RSNet~\cite{huang2018recurrent} & \textit{P}  & 56.5
    \\
    \hline
    TangentConv~\cite{tatarchenko2018tangent} & \textit{P} & 52.6
    \\
    \hline
    PointCNN~\cite{li2018pointcnn} & \textit{P} & 57.3
    \\
    \hline
    PVCNN~\cite{liu2019point} & \textit{P} + \textit{V} & 58.98
    \\
    \hline
    ASIS~\cite{wang2019associatively} & \textit{P}  & 53.4
    \\
    \hline
    KPConv~\cite{thomas2019kpconv} & \textit{P}  & \textbf{67.1}
    \\
    \hline
    CloserLook3D~\cite{liu2020closer} & \textit{P}  & 66.7
    \\
    \hline
    \textbf{DRINet}\textbf{(ours)} & \textit{P} + \textit{V} & 66.7
    \\
    \hline
    \end{tabular}
\vspace{1px}
    \caption{Results on the Area5 of S3DIS.
    }
    \label{S3DIS_results}
    \vspace{-10px}
\end{table}
\vspace{-5px}
\section{Conclusion}
    \vspace{-5pt}
    We have proposed DRINet, a novel and flexible architecture for dual-representation point cloud learning. DRINet decouples the feature learning process as SPVFE and SVPFE. In SPVFE, DRINet generates better pointwise features with the novel multi-scale pooling layer that can aggregate features at different scales. In SVPFE, the attentive gathering is proposed to deal with feature degradation when abandoning bilinear gathering operations that introduce huge memory footprints. Experiments show that our method achieves state-of-the-art \textit{mIoU} and \textit{Accuracy} in both indoor and outdoor classification and segmentation tasks with real-time speed.

{\small
\bibliographystyle{ieee_fullname}
\bibliography{egbib}
}

\end{document}